\documentclass[journal]{IEEEtran}

\usepackage{makeidx}  % allows for indexgeneration
\usepackage{url}
\usepackage{amsmath}
\usepackage{amssymb}
\usepackage{latexsym}
\usepackage{subfigure}
\usepackage{graphicx}
\usepackage{multirow}
\usepackage{color}
\usepackage{hyperref}
\usepackage{marginnote}
\usepackage{lipsum}

\definecolor{red}{rgb}{0.1,0.1,0.8}
\definecolor{blue}{rgb}{0,0,0.8}
\definecolor{green}{rgb}{0,0.4,0}

\usepackage{graphics}
\usepackage{epsfig}
\usepackage[ruled]{algorithm2e}
\usepackage{booktabs}
\usepackage{url}
\usepackage{psfrag}

\usepackage{amsfonts}
\usepackage{amsmath, amssymb}

\usepackage{epstopdf}
\usepackage{multirow}

\begin{document}

% The responses to the reviewer comments
% -> see responses.tex
% \include{responses}

\clearpage

\twocolumn
\pagenumbering{arabic}
\setcounter{page}{1}

\title{Adversarial Attack Type I: Cheat Classifiers by Significant Changes}

\author{Sanli Tang, Xiaolin Huang, Mingjian Chen, Chengjin Sun, and~Jie Yang% <-this % stops a space
\thanks{
%Manuscript received 2016;

This work is partially supported by National Natural Science Foundation of China (NSFC, 61603248, 61572315, 6151101179), 863 Plan of China 2015AA042308, and 1000-Talent Plan (Young Program).

S. Tang, X. Huang, M. Chen, C. Sun, and J. Yang are with Institute of Image Processing and Pattern Recognition and Institute of Medical Robotics, Shanghai Jiao Tong University, Shanghai, P.R. China.
(e-mails: \{tangsanli, xiaolinhuang, w179261466, sunchengjin, jieyang\}@sjtu.edu.cn). S. Tang and X. Huang contributed equally to this work.

Corresponding authors: Jie Yang and Xiaolin Huang.}}

% The paper headers
\markboth{~~~}%
{Tang \MakeLowercase{\textit{et al.}}: Adversarial Attack Type I}

% make the title area
\maketitle

\begin{abstract}
Despite the great success of deep neural networks, the adversarial attack can cheat some well-trained classifiers by small permutations. In this paper, we propose another type of adversarial attack that can cheat classifiers by significant changes. For example, we can significantly change a face but well-trained neural networks still recognize the adversarial and the original example as the same person. Statistically, the existing adversarial attack increases Type II error and the proposed one aims at Type I error, which are hence named as Type II and Type I adversarial attack, respectively. The two types of attack are equally important but are essentially different, which are intuitively explained and numerically evaluated. To implement the proposed attack, a supervised variation autoencoder is designed and then the classifier is attacked by updating the latent variables using gradient information. {Besides, with pre-trained generative models, Type I attack on latent spaces is investigated as well.} Experimental results show that our method is practical and effective to generate Type I adversarial examples on large-scale image datasets. Most of these generated examples can pass detectors designed for defending Type II attack and the strengthening strategy is only efficient with a specific type attack, both implying that the underlying reasons for Type I and Type II attack are different.
\end{abstract}


\IEEEpeerreviewmaketitle

\section{Introduction}
\IEEEPARstart{I}{n} recent years, deep neural networks (DNN) have shown great power on image classification \cite{ResNet, AlexNet}, segmentation \cite{PSPNet}, and generation tasks \cite{goodfellow2014generative, VAE}. But many
DNN models have been found to be vulnerable to adversarial examples \cite{AdversarialExample, AERobust, Survey}, which prevents deploying DNNs in physical world, e.g., self-driving and surgical robotics. Adversarial attacks reveal inconsistency between the trained DNNs and the oracle, even when they have consensus performance on known examples. This inconsistency shows that DNNs are over-sensitive to adversarial perturbations \cite{AEFirst} such that DNNs may incorrectly alter their labels when {polluted} examples only have slight difference from correctly classified ones. In form, it can be described as the following problem,
\begin{equation}\label{typeII}
\begin{aligned}
{\rm From}  {\quad} x {\quad} & {\rm Generate} {\quad} x' = {\cal A}(x) \\
{\rm s.t.} {\quad} &f_{1}(x')\;{\neq}\;f_{1}(x)\\
& f_{2}(x')=f_{2}(x)
\end{aligned}
\end{equation}
where $f_{1}$ is the classifier to be attacked while $f_{2}$ is the attacker, which could be an oracle classifier, e.g., a group of human annotators. In some literature \cite{AEFramework}, the last requirement, i.e., $f_{2}(x)=f_{2}(x')$, is written as $d(g_2(x), g_2(x')) \leq \varepsilon$, which requires the distance of $x$ and $x'$ not exceed a threshold and naturally leads to the same prediction by $f_2$.

Study on adversarial attack is meaningful to investigate the robustness of DNNs, which is crucial for many applications. For instance, the generated examples, which are generally hard for DNNs, are very valuable for enhancing the neural networks. Moreover, the relationship between the original example $x$ and the adversarial one $x'$, as well as the generating process, {is} of great importance and may result in insightful understanding for the DNNs. Therefore, since its proposal, adversarial attack, its defense, and its understanding have attracted attention. For example, the fast gradient sign method (FGSM, \cite{AdversarialExample}) can generate adversarial examples in one step via approximately linearizing the classifier. A powerful attack algorithm with a high success rate is designed in \cite{Distillation} by simultaneously minimizing the distortion of the input and the distance between the output of DNNs and the target output.

Without loss of any generality, we  {can assume that} the original sample $x$ has a positive label, i.e., $f_1(x) = f_2(x) = 1$ in (\ref{typeII}), and regard the {samples in other classes} as negative. Then via (\ref{typeII}), we generate an example $x'$ such that $f_1(x') = -1, f_2(x') = 1$. In this sense, (\ref{typeII}) produces \emph{false negatives} and induces \emph{Type II error}. Study on generating false negatives and preventing Type II error is very important. There has been great progress in the recent years; see, e.g., \cite{AdversarialExample}, \cite{AERobust},  \cite{AEFramework}, \cite{CW}, \cite{EAD}, \cite{PGD}, {\cite{unrestricted}}. However, Type II error is only one side of a coin. Partially emphasizing (\ref{typeII}) cannot lead to the ideal: a constant function $f_1(x) = 1, \forall x$ is immune to adversarial attack (\ref{typeII}) with ratio of false negatives being zero, but this function is {obviously not good}. Statistically, Type I error, corresponding to \emph{false positives}, should be simultaneously considered together with Type II error. Their cooperation gives a comprehensive measurement for a classifier: in the extreme case mentioned above, a constant function has no Type II error but has a very high ratio for Type I error, and thus it is not a good classifier.

\begin{figure}[htbp]
  \centering
    \includegraphics[width=2.5in]{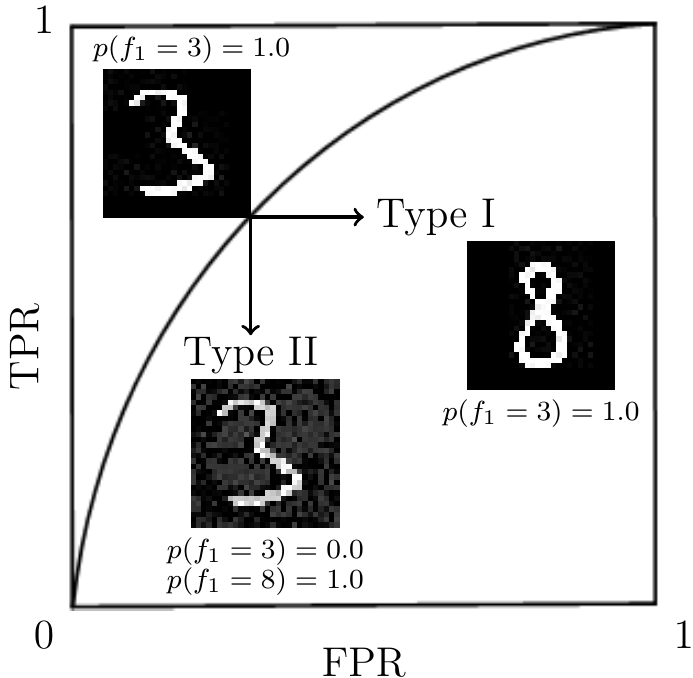}
  \caption{Relationship between Type I and Type II adversarial attacks on ROC curve of $f_1$. Through viewing number ``3'' as true sample and others as false samples, Type II attack aims to decrease the true positive rate (TPR), while Type I attack tries to increase the false positive rate (FPR)}
  \label{AdversarialFig} %% label for entire figure
\end{figure}

Adversarial {attacks} (\ref{typeII}), which {are} to increase Type II error by generating false negatives, have attracted increasing attention recently. But the discussion for Type I error is missing. For Type I error, we focus on \emph{false positives}, i.e., from an example $x$ with $f_1(x) = f_2(x) = 1$, we generate an adversarial example $x'$ such that $f_1(x') = 1$ but $f_2(x') = -1$. Mathematically,
\begin{equation}\label{typeI}
\begin{aligned}
{\rm From}  {\quad} x {\quad} & {\rm Generate} {\quad} x' = {\cal A}(x) \\
{\rm s.t.} {\quad} & f_{1}(x')\;{=}\;f_{1}(x)\\
& f_{2}(x') \neq f_{2}(x).
\end{aligned}
\end{equation}
Adversarial attack relies on inconsistency between the attacked classifier and the attacker. For (\ref{typeII}), a small change from $x$ to $x'$ that has no difference in the view of $f_2$ leads to a sign reversal of $f_1$. While the new attack  (\ref{typeI}) shows another inconsistency: $f_1$ has no response to significant changes happened in $f_2$. An intuitive example is given in Fig.\ref{AdversarialFig}, where a classifier trained on MNIST with $98.64\%$ test accuracy is attacked. Starting from a ``3'' that is correctly identified, i) by (\ref{typeII}), we slightly disturb it such that the adversarial digit is still ``3'' but is identified as ``8''; ii) by (\ref{typeI}), we really change it to an ``8'' but $f_1$ remains as before: the classifier thinks of the adversarial digit as ``3''.

The existing attack (\ref{typeII}) and the proposed one (\ref{typeI}) aim at Type II and Type I error and thus they are called \emph{adversarial attack Type II} and \emph{adversarial attack Type I}, respectively. They are equally important and irreplaceable. One may argue that Type I and Type II attack can be implemented by each other. For example, one can generate an ``8'' that is incorrectly identified as ``3'' from a correct ``8'' by the existing Type II attack methods, which however requires another example and hence violates (\ref{typeI}).
%In fact, generating an incorrect `8'' from ``3'' by Type I attack and generating one from ``8'' by Type II attack is essentially different. The former result remains properties from the original ``3'', while the later one is similar to the additional ``8''. Moreover,
The underlying reasons and the corresponding defects of the classifier for Type I and Type II attack are different, which will be detailed in Section II. As a result, many defense methods designed for the existing Type II attack do not work for the proposal Type I attack, which will be shown in Section IV.

Different to Type II attack, where the variant from the original example is quite small, Type I attack requires significant change to transform the true label. Generally, it is more hard since adding noise does not work for Type I attack. In this paper, we design a supervised variational autoencoder (SVAE) model, which is the supervised extension from original variational autoencoder (VAE) \cite{VAE}. Based on prior knowledge on Gaussian distribution, embedding label information in {the} latent space can capture the features. Then on the latent space, we attack the classifier by updating the latent variables by {a} gradient descent method, which {forwards} propagates through a decoder to recover the revised latent variables into images. Benefiting from the Gaussian restriction in the latent space and inspired by adversarial autoencoder \cite{AAE}, a discriminator is added to estimate the distribution of the manifold in the latent space and then {to} successfully implement Type I attack. Note that the Type I attack is not restricted to VAE architecture. Any framework that can transfer the input to a controllable latent space, which means that the samples are expected to follow a known distribution after coding, is possible to conduct Type I attack. {In this paper, experiments on AC-GAN \cite{ACGAN} and StyleGAN \cite{StyleGAN} for Type I attack on latent spaces (attack on latent space is designed in \cite{unrestricted} for Type II attack) are provided.} Other auto-encoder methods and generative models, including \cite{goodfellow2014generative} \cite{baluja2017adversarial} \cite{lee2017generative} \cite{xiao2018generating} \cite{huang2018introvae}, are also promising for generating Type I adversarial examples.

%In this paper, we are going to recover sparse signals from noise-corrupted measurements with saturation. The major contributions include:
%\begin{itemize}
%\item propose M1bit-CS models;
%
%\item establish the corresponding alternating direction method of multipliers (ADMM);
%
%\item design an iterative saturation detection scheme;
%
%\item evaluate M1bit-CS on one-dimensional signal recovery problems;
%
%\item apply M1bit-CS on overexpose correction for C-arm CT and achieve very promising performance.
%\end{itemize}

%and the proposed methods are then evaluated by numerical experiments. Finally, we will apply M1bit-CS to C-arm CT reconstruction for overexposure correction.

The rest of this paper is organized as follows. In Section~\ref{sec:def}, Type I adversarial attack is proposed and its link to Type II attack is discussed. The techniques of Type I adversarial attack are given in Section~\ref{sec:method}. Section~\ref{sec:numerical} evaluates the proposed attack on digits and face recognition tasks. The performance under defence designed for Type II attack is also reported, showing essential difference between Type I and Type II attack. In Section \ref{sec:conclusion}, a conclusion is given to end this paper.

%%%%%  ÊµÑéµÄµØ·½£¬»¹ÊÇ¿ÉÒÔÌáÒ»¾ä£¬ÎÒÃÇÕâ¸ö·½·¨Ò²¿ÉÒÔÓÃÀ´ Type II.  Notice that for understanding and analyzing a neural networks, the generating process  and the relationship between $x$ and $x'$ are very important and hence generating an example from other example is

\section{Type I Attack and its Relationship to Type II}\label{sec:def}

%\subsection{Type I Attack: Generate False Positives}

Deep neural networks have shown powerful fitting capability on training set. Although $f_1$ and $f_2$ have very similar performance on known data, they are still different and {the} inconsistency could be exposed by adversarial attacks. For a training example $x$, $f_1$ and $f_2$ have the same sign, e.g., $f_1(x) = f_2(x) = 1$. The inconsistency happens when $x$ becomes $x'$ in the following two scenarios: i) the attacker $f_2$ remains the same but the attacked classifier $f_1$ is over-sensitive, i.e., $f_2(x') = 1$ but $f_1(x') = -1$; ii) $f_2$ has observed the difference but $f_1$ is over-stable, i.e., $f_2(x') = -1$ but $f_1(x') = 1$.

These two types of adversarial attacks {are} described by (\ref{typeII}) and (\ref{typeI}), respectively. Type II attack (\ref{typeII}) has been insightfully investigated for attack methods, defense strategies, and theoretical analysis. But Type I attack has not been seriously considered. As summarized in \cite{Survey}, until now, only \cite{EasyFool} considers Type I error, however, it is still based on small variations and can be easily discovered. In this paper, we will design a practical Type I attack by designing a supervised variational autoencoder and by gradient updating.

\subsection{Toy Example on Feature Interpretation}
\begin{figure}[htbp]
  \centering
    \includegraphics[width=2.5in]{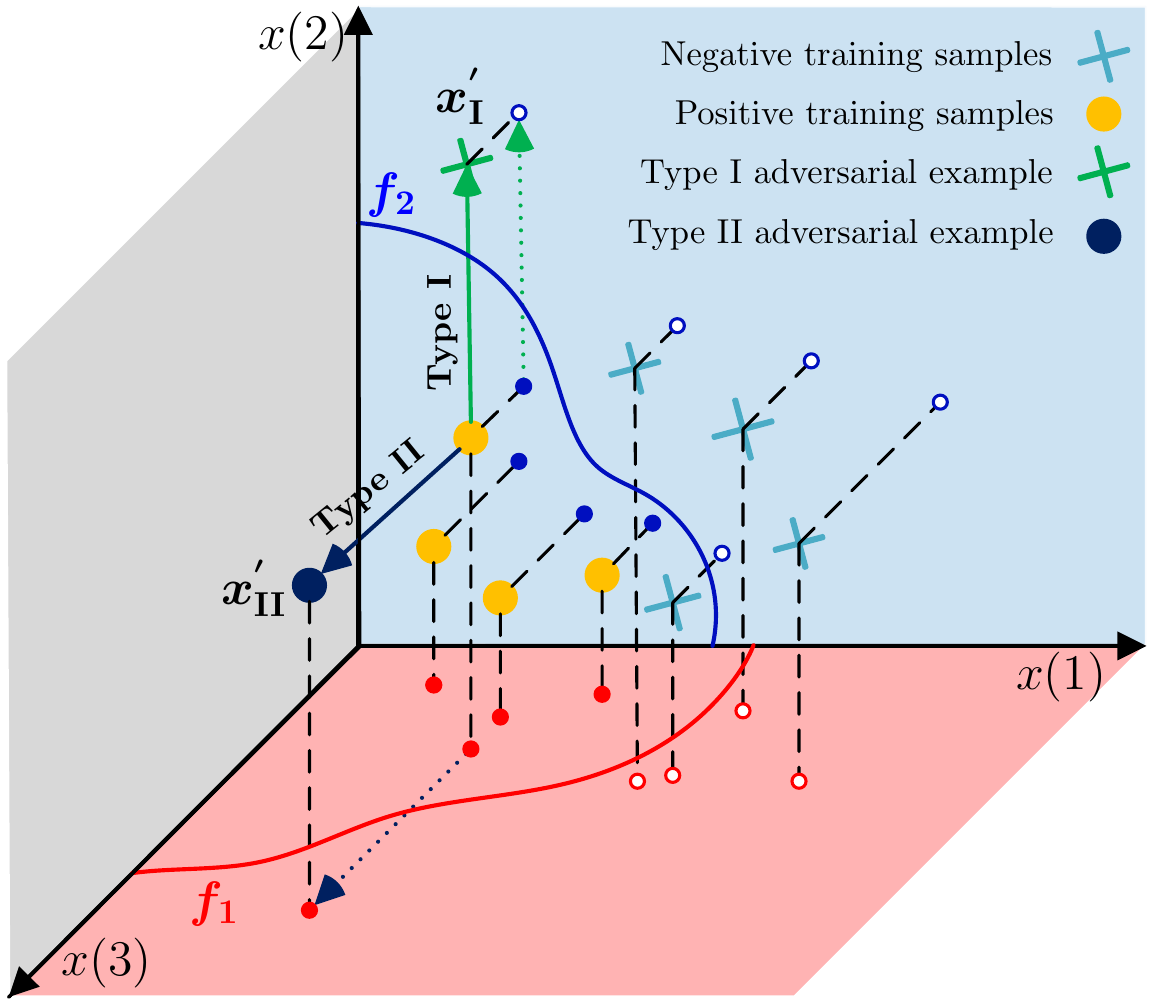}
  \caption{$f_2$ is the oracle classifier which uses $x(1)$ and $x(2)$ to distinguish samples shown by green crosses and yellow circles. But
trained on these samples, $f_1$ gets $100\%$ accuracy by considering $x(1)$ and $x(3)$.
Here $x(3)$ is an unnecessary feature such that shifting a true positive along $x(3)$
cannot be observed by {the} oracle but makes the result of $f_1$ change (Type II attack, the
blue arrow). For Type I attack (the green arrow), the sample is moved along $x(2)$
(the missing feature) and an example different to $x$ in the view of {the} oracle but $f_1(x) = f_1(x')$ is generated.}
  \label{toyexample} %% label for entire figure
\end{figure}

Before presenting the detailed technique, we here give a toy example to demonstrate different underlying reasons for Type I and Type II attack in Fig.~\ref{toyexample}, where yellow circles and green crosses stand for positive and negative training samples. The data are in {a} 3-dimensional space and the oracle classifier $f_2$ uses $x(1)$ and $x(2)$ to distinguish samples. But trained on these samples, $f_1$ gets $100\%$ accuracy by considering $x(1)$ and $x(3)$. Thus, $x(3)$ is an \emph{unnecessary feature} that is not considered by the oracle but is used in the classifier. Since $x(3)$ is an unnecessary feature, we can shift a true positive along $x(3)$ and the variant cannot be observed by the oracle but makes the result of $f_1$ change (Type II attack, the blue arrow). The link between unnecessary features and Type II adversarial attack has been theoretically discussed in \cite{AEFramework}.

To generate another type of adversarial example described in (\ref{typeI}), we need to attack \emph{missing features} instead of unnecessary ones. In Fig.~\ref{toyexample}, $x(2)$ is taken into account in the oracle but is omitted by $f_1$. Since $x(2)$ is missing, we can change the sample along $x(2)$ until it crosses the decision boundary of $f_2$ but has no influence on $f_1$. The change is displayed by the green arrow, by which we successfully generate an example different to $x$ in the view of the oracle but $f_1$ keeps unchanged. This toy example shows different essence for Type I and Type II attack. To pursue an ideal classifier that can imitate the oracle, it is required to study both unnecessary features by Type II and missing features by Type I attack.

Real applications are much more complicated than the toy example in Fig.~\ref{toyexample}, mainly because the features are not strictly orthogonal to each other. But the above feature space interpretation implies that the underlying reasons for Type I and Type II adversarial attacks are different, which follows that the defense methods designed for one type may not work for the other. In numerical experiments, we will show that some defense methods designed for Type II attack do not help the robustness to Type I attack. Worse still, if some method aims at reducing unnecessary features for defending Type II attack, it may make the classifier easier to be attacked by Type I.

Summarizingly, the existing Type II adversarial attack can cheat a classifier by small perturbations that make the classifier have undue change. The proposed Type I adversarial attack cheats a classifier by significant change that is inappropriately ignored by the classifier. The adversarial examples and the generating process for Type II have been found to be interesting and important to analyze neural networks. {For most of these topics,} Type I attack is indispensable. {Take classifier evaluation based on adversarial attacks as an example, where} not only Type II error but also Type I error needs to be considered. {Another case is classifier strengthening by re-training, for which adversarial examples generated by Type I attack, like $x'_{\bf{I}}$ in Fig.~\ref{toyexample}, cannot be found by Type II attack}. For neural network interpretation, investigating Type II adversarial attack leads to unnecessary features, but missing features can only be captured by Type I attack. {Generally, studying} adversarial examples from both Type I and Type II attack is helpful.

%The phenomena shown by Type I and Type II adversarial are
%
%Type I attack can be used in many important applications involved Type II attack.
%%The use of Type I attack is indispensable.
%\textbf{(a)} Classifier evaluation:
%%some classifiers only reveal weakness under Type I attack, e.g., a weak one considering only $x(1)$ in Fig.1 is robust to Type II attack.
%though a classifier, e.g., the one considering only $x(1)$ in Fig.1, is robust to Type II attack, it can still be Type I attacked.
%\textbf{(b)} Classifier re-training: adversarial examples that bring in new features can only be generated by Type I attack, e.g., $x'_{\textbf{I}}$ in Fig.1. \textbf{(c)} Adversarial feature generation: inconsistency between trained classifier and oracle contains both unnecessary and missing features. Unnecessary features can be detected by adversarial feature selection by Type II attack. However, it cannot find missing features, which can only be captured by Type I attack.

%\subsection{Difference between Type I and Type II Attack}

%\medskip
%{\emph{Remark: without knowing $\Psi$ makes RDCS not applicable for CT reconstruction. The above ISD scheme cannot be used for RDCS, since its hard constraints make that there is no change on $\Psi$ at the first iteration.}}

%%% add in response v1

\subsection{Toy Example on Sphere}
To investigate adversarial attacks, an interesting example to classify data from two concentric spheres with different radiuses is designed by \cite{SphereDataset}. For this simple task, the following single hidden layer network with a quadratic activation function achieves very high accuracy:
%We begin to investigate type I attack on a concentric spheres dataset \cite{SphereDataset} with equal probability assigned to each norm, where $x \in \mathbb{R}^n$ and $||x||_2$ is either 1.0 or $R $ (0.8 in our setting). Following \cite{SphereDataset}, a single hidden layer network with a quadratic activation function is deployed:
\begin{equation} \label{EqQuadraticNet}
f_1(x)=w{\bf{1}}{\cdot}(Wx)^2+b,
\end{equation}
where { $x \in \mathbb{R}^n$ and } ${W}\in\mathbb{R}^{h{\times}n}$. According to \cite{SphereDataset}, the above network is equivalent to
\begin{equation}
f_1(x)=\sum_{i=1}^d\nolimits{\alpha_i}{z_i}^2-1,
\end{equation}
where ${z}$ is a rotation of ${x}$ and $d$ is the rank of ${W}$. Based on this equivalence, it was claimed in \cite{SphereDataset} that adversarial examples could be easily found when $h > n$. Their conclusion is restricted to Type II adversarial examples. In fact, when $h < n$, there still exist Type I adversarial examples. An intuitive explanation is that when $h < n$, $n-h$ dimensions of the input are missed by the classifier such that the values on those dimensions can be arbitrarily modified while keeping the classification result of $f_1$ unchanged.

\begin{figure}
\centering
\includegraphics[width=5.4cm, height=4.0cm]{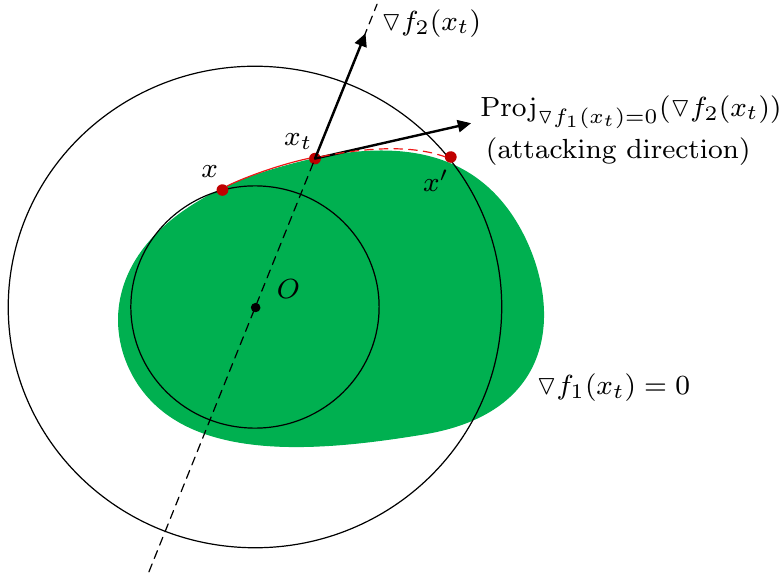}
\caption{Illustration of Type I adversarial attack on the sphere dataset \cite{SphereDataset}. The attacking direction is project of the radial vector onto the surface ${\triangledown}f_1(x_t)=0$, where $x_t$ is the sample generated at the $t$-th step.}
\label{FigSphere}
\end{figure}

Numerically, we consider an experiment that $n = 100, h = 90$ and the samples are uniformly distributed on the concentric spheres with radii being $0.8$ and $1.0$. Then a network in form (\ref{EqQuadraticNet}) is trained through one million steps with 64 batch size, which means 64 million i.i.d. training samples are used in total. After training, 100 thousand random samples are tested and the error rate is less than $10^{-4}$, showing that we have already obtained a very good classifier. For the obtained classifier, it is {hard} to generate a Type II adversarial example, as claimed by \cite{SphereDataset}. But still, from any sample $x$ with $\|x\|_2 = 0.8$,  we can first transform it to $z$-space, set
\begin{eqnarray*}
& & z_i'=z_i, \forall i=1,2,\cdots,90, \\
& & z_{91}'=z_{92}'=\cdots=z_{100}'=\sqrt{(1-0.8^2)/10},
\end{eqnarray*}
and {obtain} $x'$ from $z'$. Since the dimension $z_{91}, \ldots, z_{100}$ are missed by $f_1$, the function value of the generated $x'$ is the same as that of $x$ but $x'$ has been moved to the outer sphere. In practice when the network structure is not clear or the (back) projection is intractable, we can move $x$ to the outer sphere by projecting a radial vector onto the surface ${\triangledown}f_1(x)=0$. The trajectory from $x$ to $x'$ is illustrated in Fig. \ref{FigSphere}, which also indicates {that} the basic strategy of Type I adversarial attack we will discuss in the following sections. %where $x_t$ is the intermediate sample at the $t-{\rm th}$ step

% figure add by tsl20190328

\section{Supervised VAE Model for Type I Attack}\label{sec:method}

Finding an adversarial example to a vulnerable classifier $f_1$ is substantially an image generation task.
Both of the two types of attacks require the generated examples to belong to the wrong classes. But for Type II adversarial attack defined in (\ref{typeII}), a violate example $x'$ could be generated by directly {manipulating} the image space according to the gradients from the $f_1(x)$: typically, $x'=x+\epsilon{\nabla}_xf_1$, which is the mainstream approach, see, e.g., \cite{AdversarialExample, DeepFool, AEPhysical}. The term $\epsilon{\nabla}_xf_1$ is usually meaningless and could be regarded as {inconspicuous} noise, which follows that noise detection or de-noising can serve as defenders for Type II attack.
Such small disturbance is effective for Type II attack, for which the corresponding adversarial example remains unchanged in the view of the attacker $f_2$. However, for Type I attack, it is required to generate an adversarial example that is different to the original one in the view of the attacker. Thus, simply considering the gradients in the image space is usually not sufficient. Instead, we need to find a latent space that can capture the features used in both $f_1$ and $f_2$. By doing so, we can modify the latent variables and avoid undesirable noise or ghost when reconstructing images from the latent space.

\begin{figure}[htbp]
  \centering
    \includegraphics[width=2in, height=1.5in]{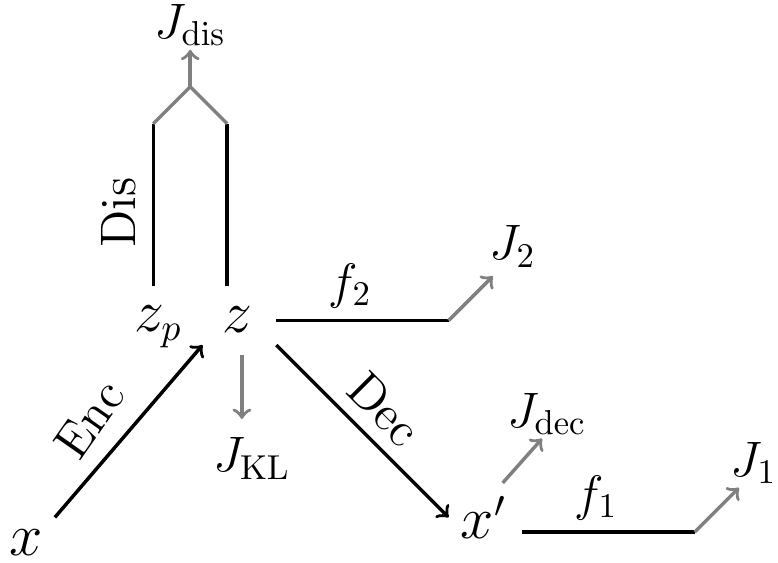}
  \caption{The framework for generating Type I adversarial example from the latent space. In this paper, SVAE is proposed for encoding.}
  \label{SVAEStructure} %% label for entire figure
\end{figure}

In Fig.~\ref{SVAEStructure}, we show the designed  framework for training the latent space and generating Type I adversarial examples. For the encoder, there are multiple possible paths and in this paper we propose a supervised variational autoencoder (SVAE), i.e., the supervised extension of VAE. For attacking,
the gradients from $f_1$ do not merely propagate to the input $x'$ of $f_1$ as the traditional methods do, but further to the latent variables $z$ through the decoder. At the same time, the gradients from the attacker are used for revising $z$ to obtain a new image $x'$ with a different label through the decoder. During this process, an equilibrium  between $f_1$ and $f_2$ is needed to keep the output of $f_1$ unchanged for achieving the Type I attack.

\subsection{Supervised Variational Autoencoder}
Let us first explain the details of SVAE.
Based on the generative model of VAE for generating $x$ in the image space from latent variables $z$, which is modeled as $p(x)={\int}p(x|z)p(z)dz$, the supervised VAE can be described as $p(x,y)={\int}p(x,y|z)p(z)dz$, where $y$ is the label information from the attacker.

Similar to VAE, in order to optimize the supervised variational autoencoder model, we find a lower bound of $p(x,y)$. Assume $q(z)$ is an arbitrary distribution in the latent space, its distance to $p(z|x,y)$ can be measured by Kullback-Leibler (KL) divergence:
\begin{equation*}\label{EqBasic}
\begin{aligned}
& {\rm KL}[q(z)||p(z|x,y)] = E_{z{\sim}q}[{\rm log}(q(z))-{\rm log}(p(z|x,y))]\\
& = E_{z{\sim}q}[{\rm log}(q(z))-{\rm log}(p(x,y|z))-{\rm log}(p(z))] \\
& ~~~ +{\rm log}(p(x,y)).\\
\end{aligned}
\end{equation*}
The above can be rewritten as:
\begin{equation*}\label{EqBasic2}
\begin{aligned}
&{\rm log}(p(x,y))-{\rm KL}[q(z)||p(z|x,y)]\\
&=-{\rm KL}[q(z)||p(z)]+E_{z{\sim}q}[{\rm log}(p(x,y|z))]\\
&=-{\rm KL}[q(z)||p(z)]+E_{z{\sim}q}[{\rm log}(p(y|z))+{\rm log}(p(x|y,z))].\\
\end{aligned}
\end{equation*}

The conditional generative methods \cite{AAE,CGAN} that use label information as a prior aim to separate the style and content in the latent space. In our setting, latent variables $z$ should contain label information as in \cite{SemiVAE}. This is mainly due to two reasons: firstly, the distribution in the latent space should be constrained by the attacker according to their labels, which enables the use of gradients from the attacker to modify the generated image; secondly, as described in \cite{AEFramework}, the judgement of the classifier is decided by a pseudometric measurement in the latent space. Therefore, we replace $p(x|y,z)$ by $p(x|z)$ since $z$ contains the label information $y$. Then, a lower bound of $p(x,y)$ can be given as:
\begin{equation}\label{EqLowerBound}
\begin{aligned}
{\rm log}(p(x,y))~{\geq} ~&-{\rm KL}[q(z)||p(z)]+E_{z{\sim}q}[{\rm log}(p(y|z))]\\
&+E_{z{\sim}q}[{\rm log}(p(x|z))].
\end{aligned}
\end{equation}

To simplify the optimization, we choose $q(z)$ to be Gaussian depending on $x$ in the latent space, which is also assumed in \cite{VAE}. Specifically, ${q(z|x)}={\mathcal{N}(\mu(x;\theta_{\rm enc}), \sigma(x;\theta_{\rm enc}))}$, from which it follows that SVAE maximizes:
\begin{equation}\label{EqSVAEObjective}
\begin{aligned}
J&~=~-{\rm KL}[q(z|x)||p(z)]+E_{z{\sim}q(z|x)}[{\rm log}(p(y|z))]\\
~&~~~ +E_{z{\sim}q(z|x)}[{\rm log}(p(x|z))] \\
~&\triangleq -(J_{{\rm KL}}+J_2+J_{dec}),
\end{aligned}
\end{equation}
where the three terms are corresponding to the encoder, the classifier, and the decoder in SVAE model, respectively. The classifier in SVAE imposes {a} restriction on latent variables to generate images with desired labels through the decoder.

Generally, a classifier trained from data only fits well on the manifold in the latent space of data but might have bad behaviour outside \cite{BadGan}. Since we directly manipulate the latent variables iteratively according to the gradients from the attacker and the attacked classifier, a discriminator is necessary to prevent the latent variables from lying outside the manifold in the latent space while attacking. Consequently, the attacker can provide {a} stable and effective direction for updating the latent variables under the restraint of the discriminator. In SVAE, the distribution of latent variables is standard Gaussian, which makes a sample-based discriminator applicable. Specifically, a binary discriminator with a sigmoid activation function on the output layer is designed to distinguish the real latent values encoded from input images by the decoder and the fake latent values sampled randomly from Gaussian distribution:

\begin{equation}\label{EqDiscriminatorObjective}
\begin{aligned}
J_{\rm dis}~=~& E_{z_{\rm true}{\sim}q(z|x)}[f_{\rm dis}(z_{\rm true})]\\
& +E_{z_{\rm fake}{\sim}\mathcal{N}(0,I)}[1-f_{\rm dis}(z_{\rm fake})].
\end{aligned}
\end{equation}

A two-stage optimization method is utilized in this paper to train SVAE. In the first stage, we simultaneously train the encoder, the decoder, and the classifier to maximize the objective function in (\ref{EqSVAEObjective}). The encoder function $f_{\rm enc}$ maps the input $x$ to a Gaussian distribution with its mean and variance to be $\mu(x;\theta_{\rm enc})$ and $\sigma(x;\theta_{\rm enc})$. By the reparameterization trick \cite{VAE}, a latent variable $z$ is sampled from such Gaussian distribution, which is then used for classification and recovery as $f_{2}$ and $f_{\rm dec}$, respectively. A gradient descent method $\Gamma$, e.g., Adam, is used for training parameters $\theta_{\rm enc}$, $\theta_{2}$, and $\theta_{\rm dec}$, which are corresponding to $f_{\rm enc}$, $f_{2}$, and $f_{\rm dec}$.

In the second stage, we train the discriminator $f_{\rm dis}$ to maximize (\ref{EqDiscriminatorObjective}) based on the well-located encoder after the first training stage. The input $x$ is first encoded into the Gaussian {with} mean $\mu(x)$ and variance $\sigma(x)$ in the latent space. Then the positives are sampled from $z_{\rm true}\sim\mathcal{N}(\mu(x;\theta_{\rm enc}),\sigma(x;\theta_{\rm enc}))$, while the negatives are sampled from the standard Gaussian $z_{\rm fake}\sim\mathcal{N}(0,I)$. A gradient descent method $\Gamma$ is applied to only update the parameters $\theta_{\rm dis}$ in the discriminator $f_{\rm dis}$. This two stage optimization algorithm is described in Algorithm~\ref{TbTrain}.

\begin{algorithm}[htbp]
\caption{Two stage learning in SVAE} \label{TbTrain}
\textbf{while} stage1Training() \textbf{do} \\
{\quad} $x,y$ $\leftarrow$ getMiniTrainingBatch()\\
{\quad}$z{\sim}\mathcal{N}(\mu(x;\theta_{\rm enc}),\sigma(x;\theta_{\rm enc}))$ \\
{\quad}$(\theta_{\rm enc},\theta_{\rm dec}, \theta_{\rm ora}){\leftarrow}(\theta_{\rm enc},\theta_{\rm dec}, \theta_{\rm ora})$  \\
{\quad\quad\quad\quad\quad\quad\quad\quad\quad}$+ {\Gamma}(\frac{\partial{J}}{\partial\theta_{\rm enc}},\frac{\partial{J}}{\partial\theta_{\rm dec}},\frac{{\partial}J}{\partial\theta_{\rm ora}})$ \\
\textbf{end while} \\
\textbf{while} stage2Traning() \textbf{do} \\
{\quad}$x,y$ $\leftarrow$ getMiniTrainingBatch() \\
{\quad}$z_{\rm true}{\sim}\mathcal{N}(\mu(x;\theta_{\rm enc}),\sigma(x;\theta_{\rm enc}))$ \\
{\quad}$z_{\rm fake}{\sim}\mathcal{N}(0,I)$ \\
{\quad}$\theta_{\rm dis}{\leftarrow}\theta_{\rm dis}+{\Gamma}(\frac{\partial{J_{\rm dis}}}{\partial\theta_{\rm dis}})$ \\
\textbf{end while}
\end{algorithm}

\subsection{Image Transition Task}

Based on the trained SVAE, we are going to generate a Type I adversarial example $x'$ from an original example $x$ as described in (\ref{typeI}) such that $x'$ and $x$ have different labels in the view of $f_2$ but are recognized as the same class by $f_1$. The first and an easier step is to transform the input image $x$ into another image $x'$ with a different label, which is an image transition task.
In our framework, the class information is not directly given, e.g., in the conditional generation methods, but comes from the supervised term in the attacker. The latent variables are revised iteratively according to the gradients from the $f_2$ and recovered into images through the decoder. Specifically, the objective function for generating targeted examples is:
\begin{equation*} \label{EqObjectiveImageTransition}
\begin{aligned}
J_{\rm IT}&=J_2(z,y')+\alpha(1-f_{\rm dis}(z))+{\gamma}||z||_2\\
&=-y'{\rm log}f_2(z)+\alpha(1-f_{\rm dis}(z))+{\gamma}||z||_2,
\end{aligned}
\end{equation*}
where the cross-entropy loss is applied to $f_2$ and it is a common choice in classification tasks \cite{ResNet, AlexNet}. Besides, a loss term with the weight $\alpha$ from discriminator $f_{\rm dis}$ is also added in order to prevent $z$ from moving outside of the manifold in the latent space. An $l_2$ regularization term on $z$ with the weight $\gamma$ is used to restrict $z$ to locate in the standard Gaussian space as the ambient space of the manifold.

\begin{figure}[htbp]
  \centering
  \subfigure[]{
    \label{fig:FigTransition:a} %% label for first subfigure
    \includegraphics[width=2.5in]{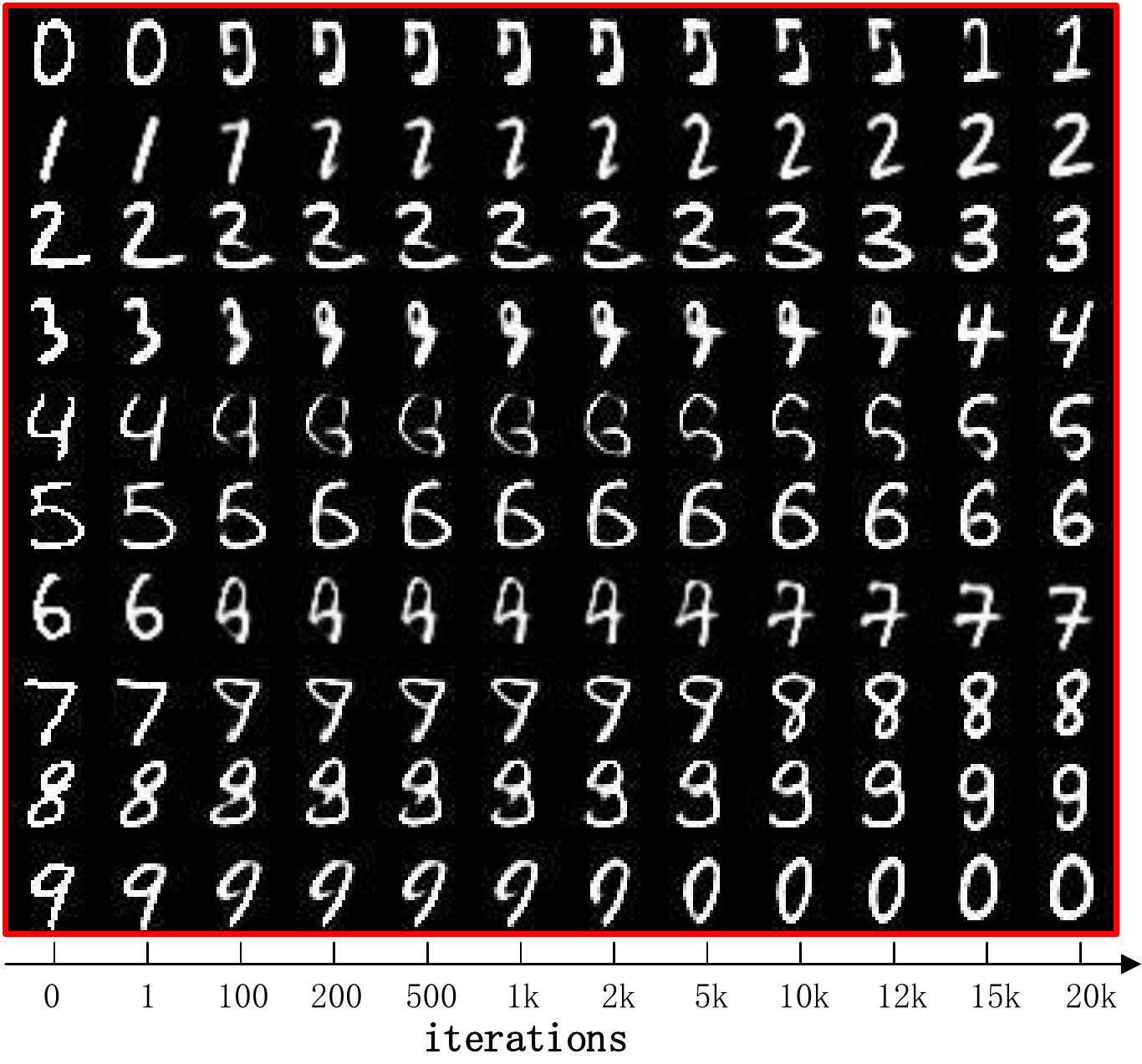}}\\
  %\hspace{0.0in}
  \subfigure[]{
    \label{fig:FigTransition:b} %% label for second subfigure
    \includegraphics[width=2.5in]{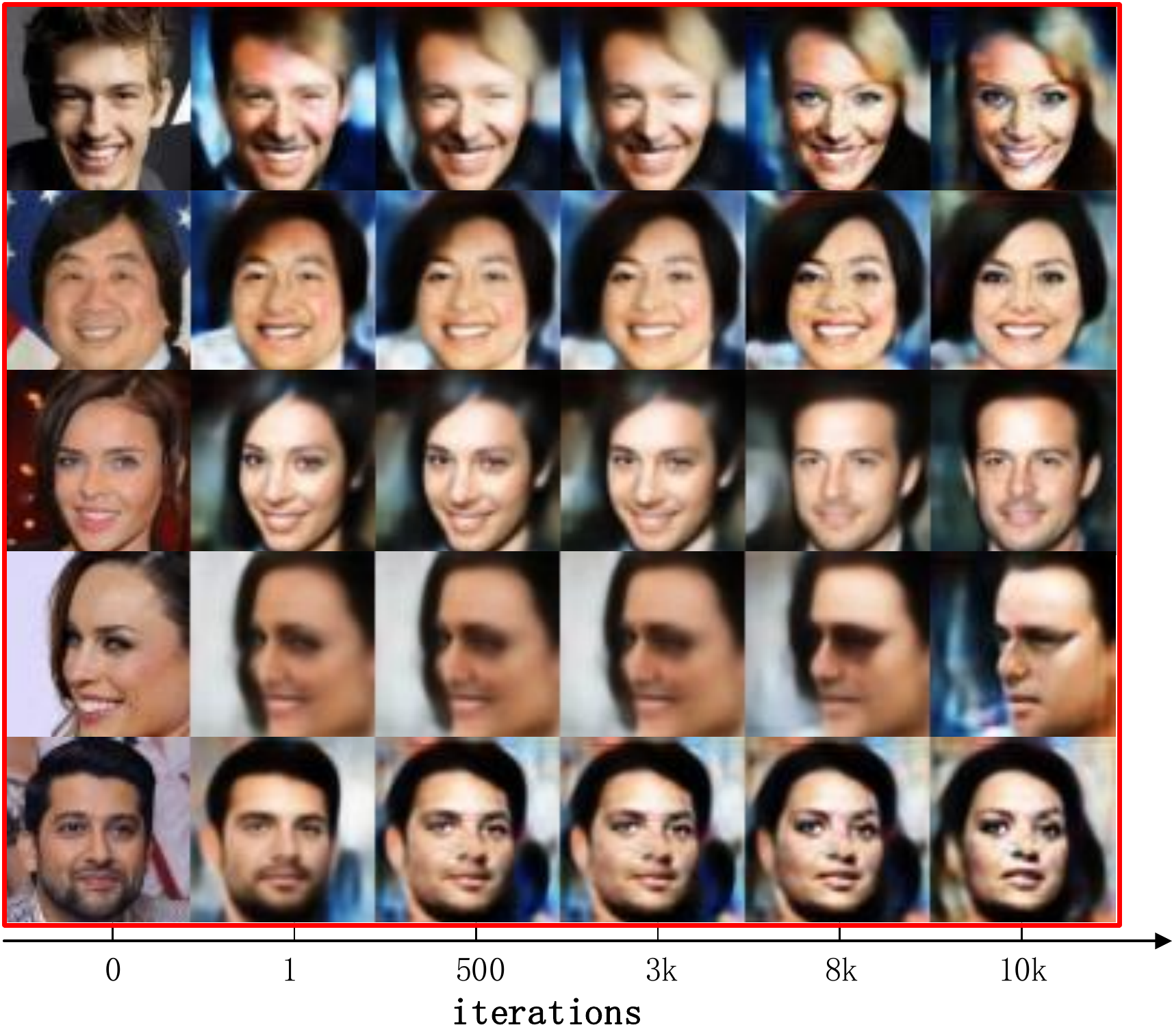}}
  \caption{Illustration of supervised image class transition by SVAE on (a) MNIST and (b) CelebA datasets. The target class for digital image $i$ is $i+1$ (the target of ``9'' is ``0'') and for male/female face images is the opposites.}
  \label{FigTransition} %% label for entire figure
\end{figure}

Here, we utilize the established method for image transition for digit images from MNIST and face images from CelebA. For MNIST dataset, a classifier with $98.64\%$ test accuracy is established and then used to image transition: from an image ``$i$'' to generate ``$i+1$''. For CelebA, a gender classifier with $94.9\%$ test accuracy is established and used to change an image's gender. The transition performance is illustrated in Fig.~\ref{FigTransition}, where the left column represents the original image, the right column shows the generated image with a desired label, and the middle ones are transitory images in different iterations.

Note that the image transition task is different to feature transformation, for which the typical route is to establish semantic features and change the required one; see, e.g., Fader Networks \cite{lample2017fader} and Deep Feature Interpolation \cite{upchurch2017deep}. The image transition here is implemented under the guide of the classifier $f_2$. For visuality, we choose $f_2$ to be a gender classifier, but it is {not necessary} to have semantic meaning and could be comprehensive. The difference {could also} be visually observed: the face generated by Fader Networks looks very similar to the original one, since the features have been decoupled; while in Fig.\ref{fig:FigTransition:b}, the generated and the original one could be regarded as different persons.

\subsection{Generating Adversarial Examples for Type I Attack}

In the above image transition task, we successfully generate new examples with desired labels. Further, incorporating the attacked classifier $f_1$ as illustrated in Fig.~\ref{SVAEStructure}, we try to keep the output of $f_1$ unchanged, i.e., to generate Type I adversarial examples as described in (\ref{typeI}).
%Based on the image transition, the attacked classifier $f_1$ can be incorporated as illustrated in Fig.~\ref{SVAEStructure} to generate Type I adversarial examples to attack $f_1$.
{For} a classifier $f_1$, an adversarial example $x'$ with {the} target label $y'$ of an input $x$ with the original label $y$ is generated through minimizing the following function, where the trained SVAE model is the attacker,

\begin{equation}\label{EqSAObjective}
\begin{aligned}
J_{\rm SA}&=J_{\rm IT}+{k_t}J_1(x',y,\cdot)\\
& = -y'{\rm log}f_2(z)+\alpha(1-f_{\rm dis}(z)) +{k_t}J_1(f_{\rm dec}(z),y,\cdot)\\
& ~~~ +{\gamma}||z||_2.
\end{aligned}
\end{equation}
Here, $J_1(x',\cdot)$ is the loss function of the attacked classifier and may take different formulations for different tasks. In multi-class classification tasks, e.g., \cite{ResNet, AlexNet}, $f_1(x)$ is a vector of probabilities for each class. While in face recognition tasks \cite{FaceNet}, $f_1(x)$ is a feature vector, where a small distance $||f_1(x')-f_1(x)||_2$ means that $x$ and $x'$ tend to be the same person. For these two tasks, the $J_1(x',\cdot)$ could be set as the following,
\begin{equation}\label{EqSAJ1}
J_{1}(x',\cdot)=\left\{
\begin{array}{clr}
-y{\rm log}f_1(x') &{\rm classification}\ \ {\rm tasks,}\\
||f_1(x')-f_1(x)||_2 &{\rm face}\ \ {\rm recognition}\ \ {\rm tasks.}\\
\end{array} \right.
\end{equation}

In (\ref{EqSAObjective}), there is a positive parameter $k_t$ reflecting the stress we put to keep the attacked classifier $f_1$ unchanged. In our method, $k_t$ could vary for different iterations. Generally, at the beginning, we allow variation on the output of $f_1$ to really change the image. Later, $k_t$ increases to pull the image back such that the output of $f_1$ is as the same as the original. Specifically, {a self-adaptive weight strategy is designed for $k_t$} to maintain such equilibrium:
\begin{equation}\label{EqKt}
\begin{aligned}
k_{t+1}~=~&k_t+{\eta}\left({\beta}J_1(f_{\rm dec}(z),y,\cdot)-J_2(z,y')\right.\\
&\left.~+{\rm max}\left\{J_1(f_{\rm dec}(z),y,\cdot)-\hat{J_1},0\right\}\right).
\end{aligned}
\end{equation}
{In our experiments, after increasing the equilibrium weight in several iterations, $k_t$ is clipped into $[0,\,0.001]$.}
The hyper-parameter $\beta$, inspired by \cite{BEGAN} to control the equilibrium between the losses $J_1$ and $J_2$, is set as
\begin{equation}\label{EqEquilibrium}
\beta=\frac{J_2(z,y')}{J_1(x',y,\cdot)}=\frac{J_2(z,y')}{J_1(f_{\rm dec}(z),y,\cdot)}.
\end{equation}
A lower $\beta$ means that the generated adversarial sample $x'$ more likely belong to the targeted label $y'$ yet the probability of $f_2$ judging that $f_2(x){\neq}f_2(x')$ is higher.
In (\ref{EqKt}), an additional term $\hat{J_1}$ is {used as} the target value for the attacked classifier. It is introduced in a hinge loss term to sacrifice a little confidence properly in $f_1$ to concentrate more on $f_2$ which {makes} the input image really {become} a new image with different classes. This term can improve the image quality and success rate for Type I attack.

We update the latent variable $z$ iteratively to minimize the objective function in (\ref{EqSAObjective}) after all those networks are well trained. At the start, we initialize $z$ to be the mean of the Gaussian $z_{\rm init}=\mu(f_{\rm enc}(x))$ according to the given input $x$. Similar to training networks, we here use Adam \cite{Adam} to optimize (\ref{EqSAObjective}) on the latent variable $z$ iteratively. The overall algorithm for generating Type I adversarial examples is illustrated in Algorithm \ref{TbAttackingMethod}. Obviously, 0 is a lower bound of $J_{\rm SA}$ in both classification and recognition tasks. Therefore, the convergence is guaranteed when optimizing $J_{\rm SA}$ through the gradient descent methods. Codes are provided in supplemental materials (SM) and will be published in the future.

\begin{algorithm}[htbp]
\caption{Generating Type I Adversarial Attack} \label{TbAttackingMethod}
given input $x$ \\
$z_0\leftarrow\mu(x;\theta_{\rm enc})$  , $k_0{\leftarrow}0$, $t{\leftarrow}0$ \\
\textbf{while} \textbf{not} converging() \textbf{do}
\quad$z_{t+1}{\leftarrow}z_{t}+{\Gamma}(\frac{\partial{J_{\rm SA}}}{\partial{z}})$ \\
\quad$x'_{t+1}{\leftarrow}f_{\rm dec}(z_{t+1})$ \\
\quad$k_{t+1}{\leftarrow}k_t+{\eta}({\beta}J_1(x'_{t+1},y,\cdot)-J_2(z,y')+$ \\
\quad\quad\quad\quad${\rm max}\{J_1(f_{\rm dec}(z),y,\cdot)-\hat{J_1},0\})$ \\
\quad$k_{t+1}{\leftarrow}{\rm max}\{0,{\rm min}\{k_{t+1},0.005\}\}$ \\
\textbf{end while} \\
$x'=f_{\rm enc}(z_T)$, $T$ is the final time step \\
\textbf{return} $x'$.
\end{algorithm}

\subsection{{Attack on Latent Space}}
{The adversarial attacks reveal weaknesses of neural networks and provide tools to analyze neural networks. Although most of the existing attacks are for image input, the attacks may happen on non-image data. Recently, an interesting attack on the latent space is designed by \cite{unrestricted}. Suppose $z$ is an original vector in the latent space and that attack is to generate $z'$ such that $\|z - z'\|$ is small but via a generative model, e.g., AC-GAN \cite{ACGAN} is used in \cite{unrestricted}, the generated images $G(z)$ and $G(z')$ are incorrectly identified by the attacked classifier $f_1$ into different classes, i.e., $f_1(G(z)) \neq f_1(G(z'))$. The original $z$ could be either given or randomly generated. This kind of attacks get rid of the encoder and are applicable for non-image data.}

{From the above description, one could observe that the attack designed in \cite{unrestricted} belongs to Type II attack. Parallelly, we could also design Type I attack on latent space, which is mathematically described as the following problem,
\begin{equation}\label{typeI-unrestr}
\begin{aligned}
{\rm From}  {\quad} z {\quad} & {\rm Generate} {\quad} z' = {\cal A}(z) \\
{\rm s.t.} {\quad} &f_{1}(G(z')) = f_{1}(G(z))\\
& \|z - z'\|_2^2 \geq \varepsilon
\end{aligned},
\end{equation}
where $f_1$ is the attacked classifier, $G$ is a generative model, and $\varepsilon$ is a user-defined threshold. Actually, (\ref{typeI-unrestr}) is a special case of (\ref{typeI}), where we use a na\"{i}ve distance-based classifier as the attacker. Basically, along a direction in the manifold where $f_{1}(G(z))$ keeps a constant, we can go away from $z$ and find Type I adversarial examples on the latent space. Specifically, those adversarial examples could be generated by minimizing the following loss function,
\begin{equation}\label{EqUnrestrictedLoss}
L(z') = k J_1(G(z'),\cdot) + \max\left\{ \varepsilon - \|z'-z\|_2^2, 0 \right \},
\end{equation}
where $J_1$ is defined as (\ref{EqSAJ1}), $G$ is a generative model that has been trained in advance, and $k$ is a trade-off parameter. The first term in (\ref{EqUnrestrictedLoss}) forces $f_1$ to judge the two generated samples $G(z')$ and $G(z)$ to be the same class, while the second term encourages $z'$ to leave the original $z$. A smaller $k$ means we allow a larger change on $f_1$ in order to obtain a larger distance in the latent space. We could design a self-adaptive strategy as (\ref{EqKt}) for $k$ but in this paper, we keep $k$ unchanged during the whole process. If an auxiliary classifier is modeled in $G$, like AC-GAN \cite{ACGAN} and is denoted by $f_G$, then we could also minimize the difference of $f_G(z')$ and $y_{\rm{target}}$ to assign a target label, otherwise, the attack is untargeted.}

%{Through optimizing $z'$ to minimize $L$, a Type I unrestricted adversarial example $x'$ will be generated by the GAN-like generator $x'=G(z')$, which will be further illustrated in our experiments.}

%GAN-like generator, e.g. the StyleGAN \cite{StyleGAN} and AC-GAN \cite{ACGAN}

\section{Experimental Validation}\label{sec:numerical}

In this section, we validate the proposed method for this new type of adversarial attack. There are two main questions: i) whether the proposed method can change an image significantly but keep the attacked classifier unchanged; ii) whether there is essential difference from the existing adversarial attack. For the two purposes, we first validate the proposed attack on MNIST dataset \cite{MNIST}, where new generated digits will be recognized as the original ones, and on CelebA dataset \cite{CelebA}, where new face images with different genders are generated but will be identified as the same person. After that, we apply defense methods to defend attacks. The different performance of Type I and Type II attack confirms their essential difference. {At last, Type I attacks on latent space are evaluated, which also shows the possibility of using different generative models for Type I attack.}

%The results show that although the defense method is efficient for Type II attack, it does not help for Type I attack, implying their essential difference.

For SVAE training, we use Adam \cite{Adam} with learning rate 0.0002. The hyper parameters $\alpha$ and $\gamma$ in (\ref{EqSAObjective}) are set to be 0.01 and 0.0001, respectively. $\beta$ for equilibrium in (\ref{EqKt}) is 0.001. During the attacking iterations, Adam is utilized to update latent variable $z$ with learning rate 0.005. The target loss $\hat{J_1}$ in (\ref{EqKt}) is set to 0.01 as the cross entropy loss for the digital image classification task and 1.00 as the Euclidean distance for the face recognition task. The detail of SVAE's architectures is provided in the SM. For datasets, MNIST contains 60K handwritten digital images of size 28x28. Following the common split, we use 50K images for training and rest for test. CelebA dataset includes more than 200K face images with 40 attribute annotations. We simply split CelebA dataset into Male/Female subsets according to their gender labels, which are then normalized and centrally cropped into size 64x64. All the experiments in this paper are implemented in Tensorflow \cite{Tensorflow} on an NVIDIA TITAN X GPU with 12GB memory.

\subsection{Type I Attack on Digits Classifier}

\begin{figure}
  \centering
  \includegraphics[width=3.0in]{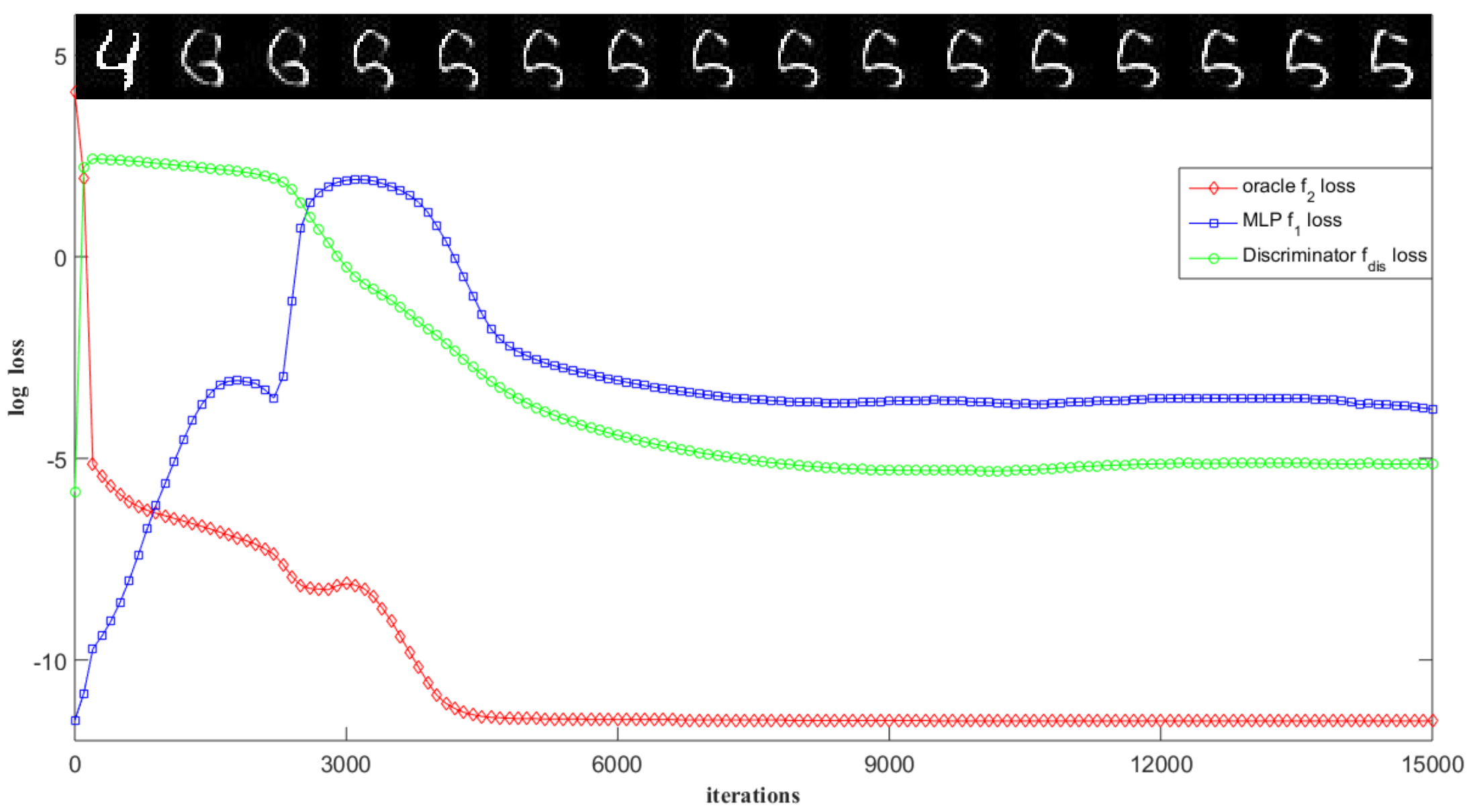}
  \caption{Illustration of the log loss terms of the attacker $f_{2}$, the discriminator $f_{\rm dis}$ and the MLP $f_{1}$ by setting $y'=5$ and $y=4$ in (\ref{EqSAObjective}). The lower loss of $f_{1}$, the more it is believe the generated image belongs to class "4". Until convergence, $f_{1}$ still classifies it as "4" with $99.41\%$ confidence.}
  \label{FigMNISTAttackLoss}
\end{figure}

First we use the proposed method to attack a classifier trained on MNIST dataset. The attacked classifier $f_1$ is an MLP with 128 hidden units and achieves $2.73\%$ test error. We train an SVAE model, which has $1.36\%$ test error and serves as the attacker $f_2$. Then from a given image that has been correctly classified by $f_1$, we apply Algorithm~\ref{TbAttackingMethod} to generate an adversarial example. The target is to change an  ``$i$'' to an ``$i+1$'' that is still wrongly classified as ``$i$'' by $f_1$.

%To validate the effectiveness of SVAE model and the algorithm for generating Type I adversarial examples in Algorithm~\ref{TbAttackingMethod}, we train SVAE model and an MLP with 128 hidden units as the attacked classifier separately on MNIST dataset. The classification error rate of the oracle classifier in SVAE and the MLP on the test set are $1.36\%$ and $2.73\%$ respectively. We here regard such MLP as a weak classifier due to its simpler architecture and higher error rate.

In Fig.~\ref{FigMNISTAttackLoss}, we illustrate the log loss terms of the attacker $f_{2}$, the discriminator $f_{\rm dis}$, and the MLP $f_{1}$, when SVAE tries to conduct a Type I attack to transform an image from ``4'' to ``5'' while the MLP still classifies it as ``4''. The images on the top are the generated adversarial examples $x'$ at the corresponding iterations. Notice that the loss for $f_2$ is relative to $y'=5$ {while the loss for $f_1$} is with respect to $y=4$. Therefore, at the beginning, the loss of $f_2$ is much higher than {that of} $f_1$ due to the fact that the original image is the number ``4''. Then, through minimizing $J_{\rm SA}$ in (\ref{EqSAObjective}), the image is gradually converted to number ``5'' while the loss of $f_1$ is increasing, since the target loss $\hat{J_1}$ is set to 0.01 and the weight for $J_1$ in (\ref{EqKt}) is clipped into 0. At the same time, the loss of the discriminator is increasing since the latent variables have to walk beyond the manifold in the latent space. When $J_1$ increases beyond $\hat{J_1}$ and $J_2$ is lower than the equilibrium described in (\ref{EqEquilibrium}), the adaptive weight $k_t$ for $J_1$ increases to pull down $J_1$ until convergence. Although the loss of the attacker $f_2$ decreases rapidly at the beginning, the supervised information from the attacker is not very reliable since its performance is unpredictable outside the manifold in the latent space. The discriminator, therefore, plays a key role in restraining the latent variables into the manifold, which makes the classifier in SVAE to provide convincible gradients.

In Fig.~\ref{FigAttackImage}(a), we show some examples of Type I attack on the digit classifier. The original images are plotted in the left column, the generated ones are in the right column, and the gradually changing process is shown between them. The number marked on the top of each image indicates the confidence of being the original class given by the attacked classifier $f_1$. During the attack, the confidence will first drop and then increase, which coincides with our setting of $k$ in Algorithm~\ref{TbAttackingMethod}, as discussed in Section III.C.

%the attacking process with the probabilities of its prediction of the MLP on MNIST dataset.

\subsection{Type I Attack on Face Recognizer}

Next, we evaluate the proposed attack on a face recognizer FaceNet \cite{FaceNet}.  We directly use the FaceNet trained on CelebA, which achieves $99.05\%$ accuracy on LFW \cite{LFW} dataset for face recognition. The same face alignment, whitening, and other pre-processing procedures are carried out as recommended \cite{FaceNet} in our experiment. An SVAE is trained on CelebA dataset with a gender classification $f_2$ of $94.9\%$ accuracy. Our task is to transform the gender of an image but keep it recognized as the same person by the FaceNet. There is an interesting difference between this task and the previous one. In the previous task, the attacker is stronger than the attacked classifier. But for this task, the SVAE classification accuracy is lower than the attacked FaceNet. Moreover, being the same gender is a necessary condition of the same person and the network of FaceNet is much deeper and more complicated. Even in the case that $f_1$ is stronger than $f_2$, the proposed Type I attack could succeed.

\begin{figure}[htbp]
  \centering
  \subfigure[]{
    \label{fig:FigAttackImage:a} %% label for first subfigure
    \includegraphics[width=3.0in]{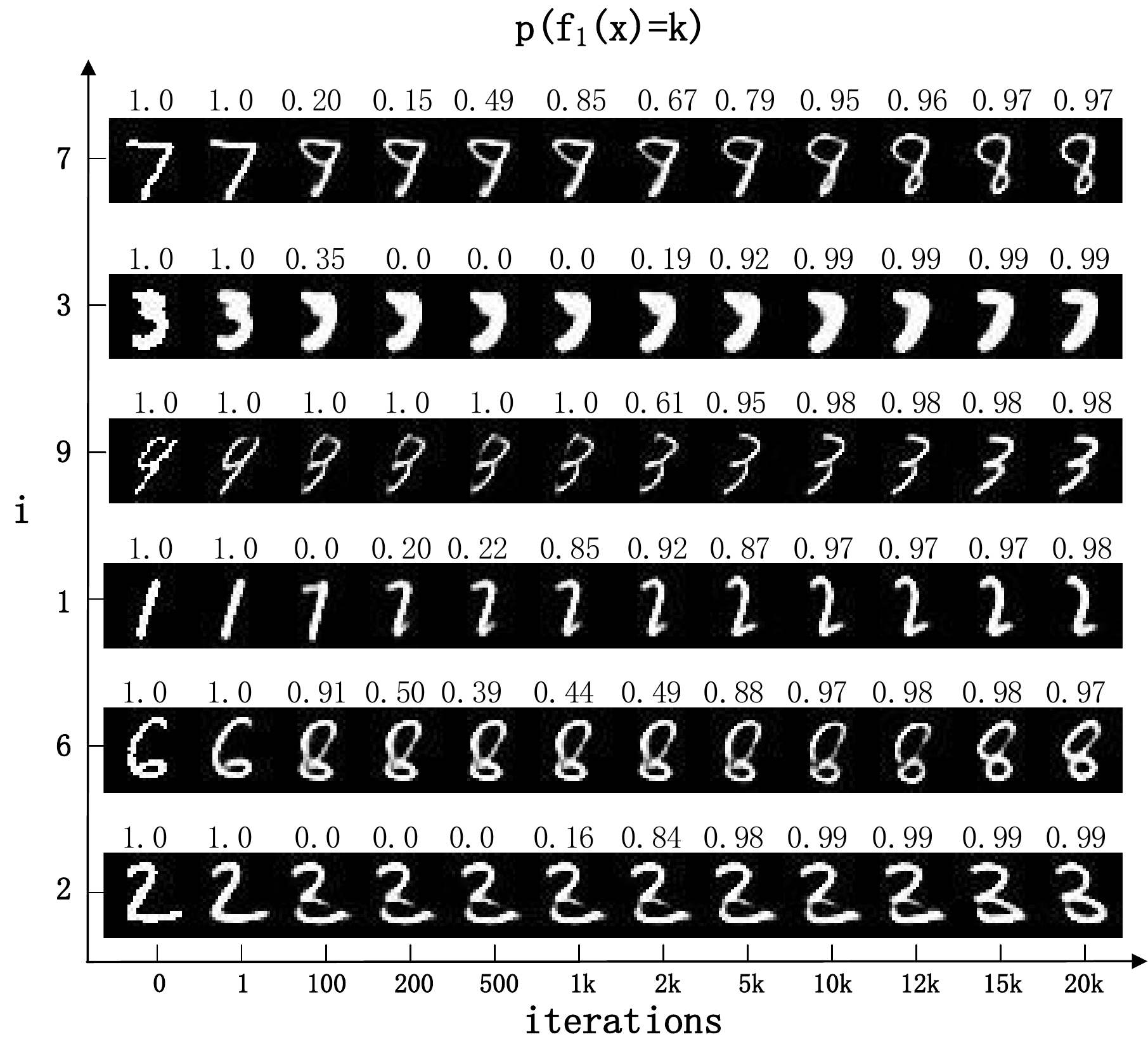}}
  \hspace{0.1in}
  \subfigure[]{
    \label{fig:FigAttackImage:b} %% label for second subfigure
    \includegraphics[width=3.0in]{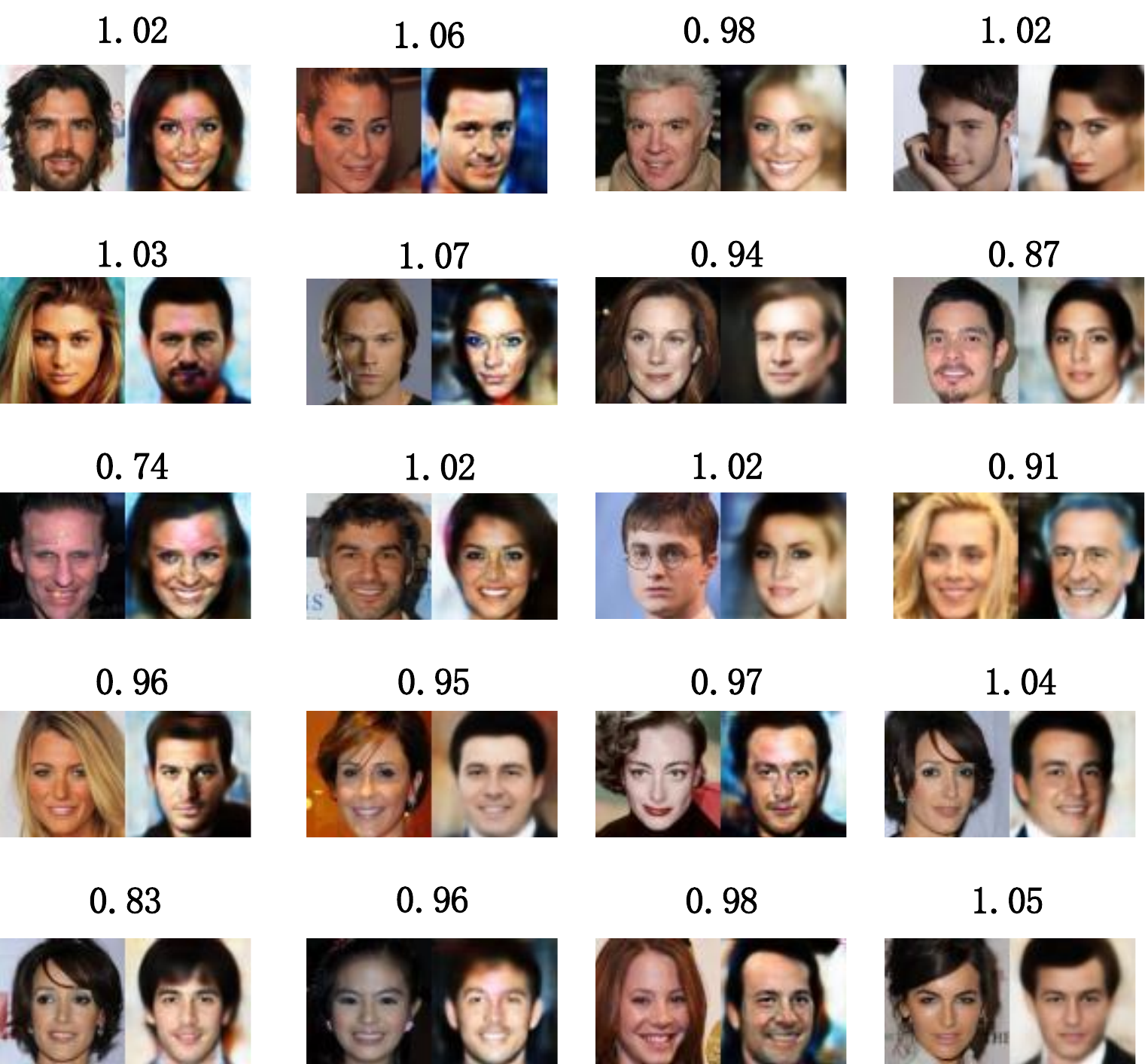}}
  \caption{(a) Adversarial examples with the probabilities according to iteration steps on MNIST dataset. (b) Adversarial example pairs with Euclidean distance on the top measured by FaceNet. Note that 1.242 is the distance threshold adopted by FaceNet to judge whether two images are from the same person.}
  \label{FigAttackImage} %% label for entire figure
\end{figure}

Some typical {adversarial} examples are given in Fig.~\ref{FigAttackImage} (b). For each pair, the left {face} is in CelebA and the right one is generated by Algorithm \ref{TbAttackingMethod}. The number above the images are the distances given by the FaceNet, for which the threshold of being the same person is suggested to be $1.242$ by \cite{FaceNet}.

%
%As recommended in FaceNet, the distance threshold is set to 1.242 \cite{FaceNet} in (\ref{EqSAJ1}). Several human experts are invited to make judgements of the last two conditions. More examples are provided in SM.
%
%The SR of Type I attack through SVAE is more than two thirds among the first thousand images on CelebA validation dataset. Some of the examples are illustrated in Fig.~\ref{FigAttackImage} (b). The results show that a weak classifier in SVAE can also conduct Type I attack against a relatively strong classifier in high success rate.

To evaluate the success rate of Type I attack, we generate adversarial examples from the first 1000 images of the validation set in CelebA. When saying that the attack is successful, we mean the image pair $(x, x')$ satisfies three conditions: i) $f_1(x) = f_1(x')$, i.e., the two faces are recognized as the same person by the FaceNet; ii) $f_2(x) \neq f_2(x')$, i.e., they are of different genders in the view of $f_2$ and are not the same person; iii) $x'$ is indeed a face. The latter two criteria are judged by thirty annotators and the average successful rate is 69.8\%. All the generated faces are provided in SM for reference.

\subsection{Type I Attack Defenses by Detecting}

The above two experiments validate the effectiveness of generating Type I adversarial examples. As previously discussed, Type II attack reflects the over-sensitivity and the Type I attack relies on the over-stability of the classifier. Since the underlying reasons are different, we expect that defense strategies designed for Type II attack by detecting the adversarial input do not help much for Type I attack.
%As discussed in Section \ref{sec:def}, since the causes for Type I and Type II attack are essentially different that Type II attack modifies the unnecessary feature while Type I attack changes the missing features both in latent space, Type I adversarial examples can still cheat the classifier with defense strategy for Type II attack.

To validate this corollary, we use the proposed method to attack the MLP with {a} feature squeezing defense strategy \cite{FeatureSqueezing} that is the most effective defense strategy and has shown great potential to counter many existing Type II attacks. {The} feature squeezing defense strategy is to detect adversarial examples by reducing the degree of input features available to adversarial examples. Specifically, it calculates the distance between the classifiers' prediction (e.g., the output of the softmax layer) of the input sample and its squeezing one. If the distance is larger than a given threshold, the defense method regards the input sample as an adversarial example and refuses to classify this sample. The defense strategy can be summarized as bellow:
\begin{equation*}\label{FeatureSqueeze}
\widetilde{f}_1(x)=\left\{
\begin{array}{clr}
f_1(x), & D(f_1(x), f_1(F(x)))< {\zeta}, \\
{\rm reject,} & {\rm otherwise,} \\
\end{array} \right.
\end{equation*}
where $F(x)$ is the feature squeezing strategy, $D(\cdot, \cdot)$ is a distance metric, and {$\zeta$ is the hyper-parameter serving as a threshold to judge whether the input $x$ is an adversarial example (then reject it) or not}.
Following the common setting in \cite{FeatureSqueezing}, the joint squeezers with one bit depth reduction and {a} $2 \times 2$ median smoothing are set for {the} feature squeezing strategy $F$, which presents the highest detection rate as revealed in \cite{FeatureSqueezing}. The distance metric $D$ is chosen to be Euclidean distance. We randomly choose 500 samples on MNIST test set and calculate the detection rate on {clean test samples, their Type I adversarial examples, and their Type II adversarial examples} with different thresholds. Type I and Type II  adversarial examples are generated by Algorithm \ref{TbAttackingMethod} (SVAE), FGSM \cite{AdversarialExample}, DeepFool \cite{DeepFool}, CW \cite{CW}, and EAD \cite{EAD}, respectively. The detection rates, i.e., the rate of being recognized as adversarial examples by the detector, for the clean samples, Type I adversarial examples, and Type II adversarial examples are listed in Table \ref{table:DetectionRate}. For different {$\zeta$} values, the detection rates for clean and Type I adversarial examples have no significant difference, which means that Type I adversarial examples cannot be well distinguished from clean data. By contrast, the detection ratio for Type II adversarial examples {is} significantly higher than that for clean data, showing that feature squeezing is a good detection method for Type II attack, but does not work for Type I attack.

\begin{table}[htbp]
\centering
\caption{Detection rates on MNIST test set, Type I adversarial examples and Type II adversarial examples with different thresholds.}
\label{table:DetectionRate}
\begin{tabular}{c|c|ccc}
\toprule
\multicolumn{2}{c|}{{$\zeta$}} & 0.2510 & 0.6089 & 0.9353 \\
\midrule
\multicolumn{2}{c|}{Clean} & {0.234} & {0.102} & {0.050} \\
\midrule
\multirow{4}*{Type II} & FGSM \cite{AdversarialExample} & 0.870 & 0.776 & 0.726 \\
& DeepFool \cite{DeepFool} & 0.998 & 0.974 & 0.872 \\
& CW$\_l_2$ \cite{CW} & 0.998 & 0.976 & 0.842 \\
& EAD \cite{EAD} & 0.998 & 0.980 & 0.902 \\
\midrule
Type I & SVAE & {0.306} & {0.122} & {0.076} \\
\toprule
\end{tabular}
\end{table}

{Similar phenomenon on the} feature squeezing could also be observed in attacking FaceNet, for which 1000 Type I adversarial examples are generated as described in Section IV.B. The same number of Type II adversarial examples are generated by {the} projected gradient descent (PGD, \cite{PGD}). We set {$\zeta=1.2$} and consider different sizes of the smoothing kernel. The detection rates can be found in Table \ref{table:TbCelebADefenseSqeeuze}, showing that feature squeezing does not help for Type I attack since Type I attack is not built on adding noise.

\begin{table}[htbp]
\centering
\caption{Detection rates on CelebA test set through the feature squeezing defense method.}
\label{table:TbCelebADefenseSqeeuze}
\begin{tabular}{c|c|cc}
\toprule
Filter size & Clean & Type I (SVAE) & Type II (PGD) \\
\midrule
$3{\times}3$ & 0.000 & {0.000} & 0.487 \\
$5{\times}5$ & 0.064 & {0.032} & 0.971 \\
\bottomrule
\end{tabular}
\end{table}

\subsection{Type I Attack Defenses by Strengthening}
%{
%Feeding back the adversarial examples to the neural network during training is another method to increase the robustness of networks. This natural strategy is valid for both Type I and Type II attacks. But some advanced techniques, like Adversarial Logit Pairing \cite{LogitPairing}, which are designed for Type II adversarial examples, will lose effect on Type I attack. The key idea of Adversarial Logit Pairing is to a pairwise loss $L(f(x),f(x'))$, which is typically a distance measure between the original example $x$ and the adversarial one $x'$. By minimizing this loss, we encourage the logits between $x$ and $x'$ to be similar. Clearly, this setting works only for Type II attack. If $x'$ is generated by Type I attack, then $x$ and $x'$ should not belong to the same class and the logit pairing technique cannot be directly used with the Type I adversarial examples.}

Feeding back the adversarial examples to
{re-train the classifier} is another method to increase the robustness of networks.
In the following, we use MNIST dataset to test the defending performance of Adversarial Logit Pairing \cite{LogitPairing}, which is designed for Type II adversarial examples.
The attacked classifier is LeNet \cite{LeNet}. After strengthening, adversarial examples generated by attacking the original networks are used for validation. The accuracy for Type II attack is improved to $98.5\%$, since minimizing the pairwise logit pairing loss $L(f(x),f(x'))$ that requires a small distance between the original example $x$ and the adversarial one $x'$ is helpful. However, this setting does not work for Type I attack: if $x'$ is generated by Type I attack, then $x$ and $x'$ do not belong to the same class and $L(f(x),f(x'))$ is not expected to be small. In our experiment, the accuracy for Type I attack is only $14.2\%$, verifying that the logit pairing technique cannot be directly used for the Type I adversarial examples.

As suggested and numerically verified by \cite{Ensemble}, adding adversarial examples generated from other attacks is helpful, which is based on the transferability across different models and attack methods. An interesting question is that whether the capability of defending one type of adversarial attack could help against another type. To answer this question, we strengthen FaceNet by Type I, Type II (FGSM), and Type II (PGD) adversarial examples, respectively. For Type II attack, we use the adversarial logit pairing method. For Type I attack, the adversarial logit pairing does not help as shown before, so we turn to vanilla adversarial training. Specifically, adversarial examples generated by attacking the original networks are randomly split into two parts with equal probability. One subset is sent back together with LFW dataset for adversarial training. Then strengthened FaceNets are evaluated on the rest adversarial examples. The classification accuracy could be found in Table \ref{table::ensemble}. The diagonal shows the performance of defending the adversarial attack by adversarial training on the same attack method. Other elements show the transferability of adversarial training. One could find that strengthening FaceNet by FGSM can improve the robustness against PGD; and vice versa.
However, strengthening by Type II adversarial examples has little effect on defending Type I attack, showing their difference.
%At least in this experiment, adversarial training by Type I adversarial examples seems helpful for Type II attack as well, which is beyond our expectation but is of interest for further study.
%one could find that strengthening FaceNets by Type II adversarial examples have little effect on defending Type I adversarial attack.  But the results implies that adversarial training by Type I adversarial examples seems helpful for Type II attack as well, at least in this example, which is beyond our expectation but is of interest for further study. }
\begin{table}[htbp]
\centering
% caption is edited by tsl20190328
\caption{Classification accuracy of FaceNet strengthened by different attack methods on adversarial examples}
\label{table::ensemble}
\begin{tabular}{l|c|ccc}
\toprule
  &       & Type I  & Type II  & Type II \\
\emph{strengthened} & LFW & (SVAE)  & (FGSM)    & (PGD) \\
\midrule
%FaceNet   & - & - & - \\
\emph{by} SVAE  & 0.990 & 0.554 & 0.987 & 0.998 \\
\emph{by} FGSM  & 0.995 & 0.110 & 1.000 & 0.996 \\
\emph{by} PGD   & 0.997 & 0.140 & 0.998 & 1.000 \\
\bottomrule
\end{tabular}
\end{table}

%}

{\subsection{Type I Attack on Latent Space}}
{In the last experiment, we will consider Type I attack on latent spaces. Attacks could happen on non-image data. But to visualize the performance, we apply a generative model to transfer a latent vector to an image. The attack design is independent of the generative model. For attacking LeNet \cite{LeNet} on MNIST, we choose AC-GAN \cite{ACGAN}, which is also used in \cite{unrestricted}. For attacking FaceNet on CelebA \cite{CelebA}, we use StyleGAN \cite{StyleGAN}, which is one of the state-of-the-art generative models. We train an AC-GAN with 128 latent features on MNIST and use StyleGAN with 512 latent features provided by \cite{StyleGAN}. Since these generative models only affect the visual quality but have no difference on attack performance, we do not report their structures and training details here. Based on trained generative models, we attack the classifier $f_1$ by minimizing (\ref{EqUnrestrictedLoss}), where the hyper-parameters are set as ${k}=10^{-2},\varepsilon=0.1$ for attacking LeNet, and ${k}=10^{-3},\varepsilon=0.35$ for attacking FaceNet, respectively. Adam with the learning rate $0.01$ is used.}

\begin{figure}[htbp]
  \centering
    \includegraphics[width=3.6in]{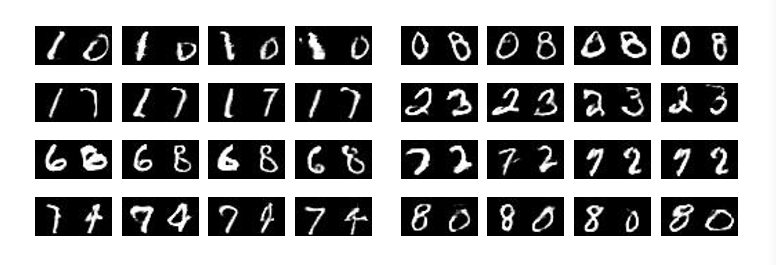}
  \caption{{Type I attack performance on the latent space of AC-GAN. For each pair of the digits, the attacked LeNet correctly recognizes the left digit but incorrectly thinks the right one belongs to the same class.}}
  \label{FigUnrestricted-mnist} %% label for entire figure
\end{figure}

{In Fig. \ref{FigUnrestricted-mnist}, Type I attack on latent space is displayed by plotting the generative images (in $28 \times 28$) by AC-GAN. Each pair of those digits are visually different, however, they are identified to the same class by the attacked LeNet. Since AC-GAN has an auxiliary classifier, we can utilize this classifier to set a target label.}

%{In Fig. \ref{FigUnrestricted}, we show the Type I unrestricted adversarial examples of human face and digit images. With the help StyleGAN and its impressive generative ability, the adversarial human face images are generated in size of $1024{\times}1024$, which is also of interests for future study on investigating the Type I attack with other GAN-like generative models.}

\begin{figure}[htbp]
  \centering
    \subfigure{
    \includegraphics[width=3.3in]{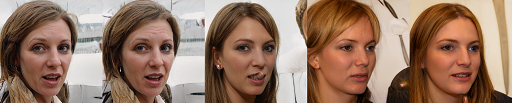}}
    \subfigure{
    \includegraphics[width=3.3in]{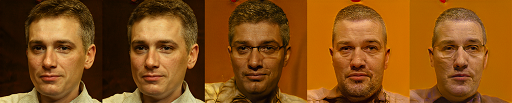}}
    \subfigure{
    \includegraphics[width=3.3in]{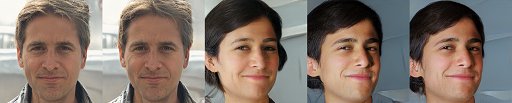}}
    \subfigure{
    \includegraphics[width=3.3in]{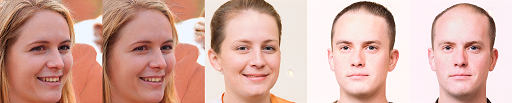}}
    \subfigure{
    \includegraphics[width=3.3in]{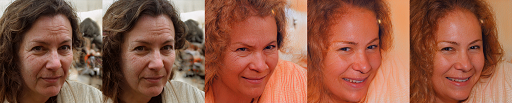}}
    \subfigure{
    \includegraphics[width=3.3in]{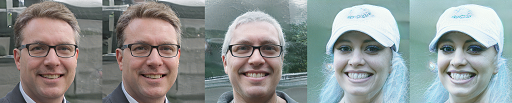}}
  \caption{{Type I attack on the latent variables of StyleGAN. The left column shows the generated faces corresponding to the starting latent variable, the right column shows the attack results, and the intermediate images are displayed in the middle. The image size is $1024 \times 1024$. For all the 6 rows, the distances judged by FaceNet between the left and the right are: 0.6780, 0.6321, 0.7632, 1.0482, 0.6448, and 1.0449 (from top to bottom), which are all below 1.242. Though the appearances have been changed significantly, the faces in each row are still identified as a same person by FaceNet.}}
  \label{FigUnrestricted-face} %% label for entire figure
\end{figure}

{In Fig. \ref{FigUnrestricted-face}, we display 6 face sequences generated by the Type I adversarial attack on the latent space of StyleGAN. Thanks to the great generative ability of StyleGAN, the adversarial images are in the size of 1024x1024. Starting from the left face in each row, the attack significantly changes the appearance. However, the attacked FaceNet still recognizes them as the same person. Differently to the attack shown in Fig.\ref{fig:FigAttackImage:b}, the attack based on StyleGAN can not control the change direction, e.g., from male to female or the opposite. In the future, it is of great interest to implement targeted Type I attack based on StyleGAN or other generative models.}

\section{Conclusion}
\label{sec:conclusion}
False positive and false negative rates are important evaluation criteria for the performance of classifiers, thus, adversarial attacks aiming at false positivists (Type I attack, proposed in this paper) and at false negatives (Type II attack, the current popular ones) are both deserved {to be investigated}. The inconsistency in the feature spaces makes classifiers vulnerable. But the underlying reasons for Type I and Type II attack are different: Type I attack relies on the missing features that are utilized by the attacker but are ignored by the attacked classifier, while Type II attack modifies the unnecessary features that are meaningless for the attacker  but are concerned by the attacked classifier.

To generate {false positives} is to cheat a classifier by significant changes.
The generative target is a totally new sample that is misclassified into the same class of the original by the attacked classifier. For Type I adversarial attack,
a supervised variational auto-encoder framework is designed as a generative method. In this framework, the attacker is modeled explicitly to provide supervised information for generating a new and meaningful adversarial example. Rather than directly manipulating in the image space which leads to noise eventually, a generating algorithm for Type I attack is then established by revising the latent variables in the latent space according to the gradients {from both} the attacker and the attacked classifier, and then those updated variables are recovered into images through a decoder. In order to obtain stable and convincible gradients from the attacker, a discriminator is designed to restrict latent variables to locate on the manifold through the iterative generating algorithm.

In numerical experiments, the proposed method successfully generates Type I adversarial examples to cheat well trained classifiers. Most of these examples can pass feature squeezing detection, an efficient detection method for Type II attack, implying the different essential reasons for Type I and Type II attack. Because of the difference, the current strengthening methods do not have type-crosses capability, i.e., retraining neural networks by Type II adversarial examples have no benefits on defending Type I attack.
Generally, Type I adversarial attack is a new adversarial attack and is of great importance for understanding neural networks.
The proposed method is only one way to generate Type I adversarial examples and other architectures with auto-encoder and generative capability are also promising for Type I attack, which can be used for classifier evaluation, classifier re-training, and feature analysis.

\section*{Acknowledgment}
The authors thank Mr. Sizhe Chen in Shanghai Jiao Tong University for helpful discussions.

%The authors are grateful to the anonymous reviewers for their insightful comments.

%The authors would like to thank...

\bibliographystyle{IEEEtran}        % Include this if you use bibtex
\bibliography{refs}

\end{document}